\documentclass{article} 
\usepackage{iclr2027_conference,times}

\usepackage{amsmath,amsfonts,bm}

\def\eqref#1{equation~\ref{#1}}

\def\1{\bm{1}}

\DeclareMathAlphabet{\mathsfit}{\encodingdefault}{\sfdefault}{m}{sl}
\SetMathAlphabet{\mathsfit}{bold}{\encodingdefault}{\sfdefault}{bx}{n}

\usepackage{hyperref}
\usepackage{url}
\usepackage{graphicx}
\usepackage{svg}
\usepackage{booktabs}
\usepackage{placeins}
\usepackage{wrapfig}
\usepackage{tabularx}
\usepackage{array}

\title{Flow-JEPA: Robust Latent Dynamics for JEPA World Models via Flow Matching}

\author{Yanchen Huo \\
Nanyang Technological University\\
\And
Ziying Song\thanks{ Use footnote for providing further information
about author (webpage, alternative address)---\emph{not} for acknowledging
funding agencies.  Funding acknowledgements go at the end of the paper.}  \\
Nanyang Technological University \\
\And
Yadan Luo \\
University of Queensland \\
}

\author{
Yanchen Huo$^{1}$ \quad
Ziying Song$^{1}$\thanks{Corresponding author} \quad
Yadan Luo$^{2}$ \\
$^{1}$Nanyang Technological University \quad
$^{2}$The University of Queensland \\
}

\iclrfinalcopy 
\begin{document}

\maketitle

\begin{abstract}

Joint-Embedding Predictive Architectures (JEPAs) provide a powerful framework for latent world modeling and planning in a reconstruction-free manner. Although numerous JEPA-based approaches have been proposed to mitigate representation collapse, our experiments on localized, out-of-distribution visual noise reveal that performance degradation remains pronounced and unresolved. We propose Flow-JEPA (F-JEPA), a flow-based latent dynamics model that jointly generates a sequence of future latent states conditioned on the current observation and actions. A Gaussian distribution serves as the flow source, exposing the vector field to perturbed latent trajectories as it learns to transport them toward clean future representations. This formulation retains the reconstruction-free JEPA framework while switching from pointwise transition regression to stochastic trajectory-level prediction. F-JEPA raises mean success from $86\%$ to $92\%$ under clean observations and from $67\%$ to $86\%$ under noisy conditions. Further evaluations over varying perturbation severity and inference settings show that the performance advantage persists across a broad range of conditions. These results suggest that conditional flow matching provides a promising alternative to deterministic autoregressive prediction as a dynamics formulation in JEPA world models. Code is available at \href{https://github.com/HuoYanchen/Flow-JEPA}{https://github.com/HuoYanchen/Flow-JEPA}.
\end{abstract}

\section{Introduction}

World models learn predictive representations of environment dynamics by modeling the future consequences of actions, providing a foundation for planning, policy learning, and synthetic data generation. Early and influential approaches learn compact latent dynamics from high-dimensional observations and use the learned model for imagination-based control or policy optimization \citep{ha2018world,hafner2019dream}. More recent systems, such as IRIS \citep{micheli2022transformers}, further demonstrate that sequence models can serve as powerful world models by combining learned visual tokenization with autoregressive dynamics prediction. Despite their success, many such models rely on pixel-level reconstruction or generative prediction objectives. Although these objectives preserve rich visual detail, they can also force the model to spend capacity on task-irrelevant factors such as background or lighting. This motivates reconstruction-free latent world models that predict future states in an abstract representation space rather than reconstructing raw pixels.

Joint-Embedding Predictive Architectures (JEPAs) \citep{jepa} provide a fundamental framework for this reconstruction-free setting. Instead of predicting pixels, JEPA-style methods encode observations into latent embeddings and train a predictor to forecast missing or future representations. LeWorldModel (LeWM) \citep{maes2026leworldmodel} extends this idea to compact, end-to-end world modeling directly from pixels in reward-free environments. It learns an action-conditioned latent dynamics predictor, while using the Sketched-Isotropic-Gaussian Regularizer (SIGReg) \citep{balestriero2025lejepa} to prevent representation collapse without reconstruction losses, frozen encoders, exponential moving averages, or auxiliary supervision. Avoiding reconstruction, however, does not by itself guarantee that the resulting dynamics models are insensitive to nuisance variation.

\begin{figure}[h]
\begin{center}
\includegraphics[width=5.5in]{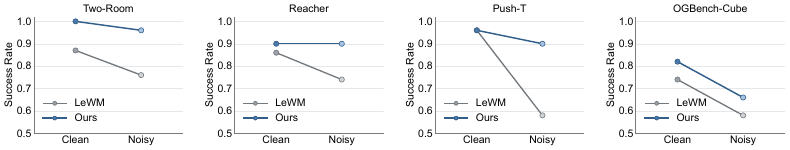}
\end{center}
\caption{\textbf{Latent-dynamics formulation affects planning robustness.}
F-JEPA, which instantiates trajectory-level vector-field dynamics,
achieves higher average success under clean observations and substantially
smaller degradation under noisy observations than the autoregressive
LeWM baseline across four environments. The $y$-axis starts at 0.5.}
\label{fig:sr}
\end{figure}

A common formulation of latent world-model dynamics is deterministic autoregressive prediction, in which a one-step state-transition model is recursively applied to generate a future trajectory. This design has two structural properties that make it vulnerable to perturbations:
\begin{itemize}
    \item \textbf{Error accumulation.} The latent history, increasingly composed of model predictions, is reused as input to subsequent transitions, allowing perturbation-induced deviations and prediction errors to propagate across the rollout.
    \item \textbf{Pointwise supervision.} The next-latent regression objective supervises the predictor at observed latent targets. While this objective encourages accurate prediction on clean training trajectories, it provides no explicit mechanism for modeling or recovering from local perturbations of the latent state.
\end{itemize}

Using LeWM as a representative autoregressive JEPA world model, we find that planning can degrade substantially under localized visual perturbations that leave the underlying physical state and environment dynamics unchanged. This failure mode suggests that robustness may depend not only on how observations are encoded, but also on how dynamics are modeled within the learned representation space, raising a broader question: \emph{to what extent does the formulation of latent dynamics determine the robustness of JEPA world models?}

We investigate trajectory-level conditional vector-field prediction as an alternative formulation of latent dynamics. The resulting model, Flow-JEPA (F-JEPA), jointly processes the complete future latent trajectory and uses a conditional flow matching (CFM) objective to learn its evolution conditioned on the current latent state and an action sequence. CFM learns a time-dependent vector field that transports samples from a source distribution toward a conditional target distribution and has been used for continuous generation in images, video, and robot actions \citep{lipman2022flow, wan2025wan, black2024pi_0}. During training, the vector-field predictor is supervised at stochastic interpolation states between source and target trajectories rather than only at clean latent endpoints.

SIGReg and CFM introduce Gaussian structure at different stages of the model. SIGReg regularizes the marginal encoder distribution, whereas CFM uses a Gaussian source to define stochastic transport paths for dynamics learning. By exposing the dynamics model to latent states beyond the observed transitions, such training may improve tolerance to perturbation-induced deviations. We test this hypothesis across four control environments. F-JEPA improves average planning performance under clean conditions and exhibits substantially smaller degradation under perturbations than the autoregressive baseline. Mean success increases from $86\%$ to $92\%$ under clean observations and from $67\%$ to $86\%$ under noisy observations. Additional stress tests show that the observed advantage is not confined to a single perturbation strength or inference configuration.

Our main findings are summarized as follows:
\begin{itemize}
    \item We identify and characterize a robustness failure mode in a representative autoregressive JEPA world model: task-irrelevant visual perturbations can substantially degrade latent-space planning even when the underlying task dynamics are unchanged.
    \item We study latent-dynamics robustness by contrasting deterministic autoregressive state-transition prediction and trajectory-level conditional vector-field prediction, and instantiate the latter formulation as F-JEPA.
    \item Across four control environments, the flow-based dynamics model increases average planning performance and substantially reduces degradation under visual perturbations. Additional analyses over varying perturbation severity and inference settings show that this trend persists across a broad range of evaluation conditions.
\end{itemize}

\section{Related Work}

\paragraph{Latent World Models.} Latent world models compress high-dimensional observations into compact state representations and model their temporal evolution for prediction and control. Early approaches typically combine latent dynamics with observation reconstruction and task-specific reward supervision \citep{ha2018world, hafner2019learning}. The Dreamer family replaces online action search with actor-critic learning over trajectories imagined by the world model \citep{hafner2019dream, hafner2020mastering, hafner2023mastering, hafner2025training}. Other generative approaches include IRIS \citep{micheli2022transformers}, which models discrete visual tokens with an autoregressive Transformer, and DIAMOND \citep{alonso2024diffusion}, which uses diffusion-based observation prediction for Atari control. Recent work has further explored stochastic generative forecasting in fixed pretrained feature spaces \citep{walker2025frozen, porcher2026flow}. Our focus instead lies in reconstruction-free JEPA world modeling, where we study how the formulation of latent predictive dynamics affects planning robustness.

\paragraph{Joint-Embedding Predictive Architectures.} Joint-Embedding Predictive Architectures (JEPAs) learn to forecast target embeddings without reconstructing raw observations. I-JEPA \citep{assran2023self} and V-JEPA \citep{bardes2024revisiting, assran2025v} show that representation prediction can learn strong image and video features without pixel reconstruction. This idea has recently been adopted for world modeling. DINO-WM \citep{zhou2024dino} and PLDM \citep{sobal2026learning} are important examples of JEPA-style world models for control. PLDM learns visual representations end-to-end, whereas DINO-WM models dynamics in a frozen pretrained feature space. Our study builds on LeWorldModel \citep{maes2026leworldmodel}, which trains a compact end-to-end JEPA world model using a next-embedding prediction objective and SIGReg \citep{balestriero2025lejepa}. F-JEPA adopts the Gaussian-regularized encoder and models latent dynamics with CFM over complete future trajectories, providing a trajectory-level vector-field alternative to autoregressive one-step prediction.

\paragraph{Flow Matching for Conditional Generation.} Flow matching is a generative modeling framework that learns a time-dependent vector field transporting samples from a source distribution to a target data distribution \citep{lipman2022flow, liu2022rectified, liu2022flow}. Compared with direct regression, flow-based objectives provide a flexible way to model continuous conditional distributions and have been applied to domains such as image generation, video generation, and robot action generation \citep{wan2025wan, black2024pi_0}. Recent work has explored replacing uninformed Gaussian initialization with structured source distributions \citep{jia2026action, kim2026better}, exploiting temporal continuity and strengthening conditional generation. Our work applies CFM to future JEPA latent states, providing a distributional formulation of latent dynamics.

\section{Rethinking Latent Dynamics in JEPA world models}
\label{sec:motivation}

\subsection{Stable Representations Do Not Imply Robust Dynamics}

Reconstruction-free prediction creates an opportunity to discard visual detail that is unnecessary for modeling environment dynamics \citep{sun2026vla}. This opportunity should not, however, be confused with a guarantee of robustness. A JEPA world model can learn a non-collapsed, well-structured representation while its action-conditioned predictor remains sensitive to changes within that representation space.

LeWM \citep{maes2026leworldmodel} provides a useful
example of this distinction. It jointly trains a visual encoder, which maps an observation $o_t$ to a latent state $z_t=f_\xi(o_t)$, and a dynamics predictor, which forecasts the next latent from the current latent and action. Because next-embedding prediction alone admits a constant collapsed solution, LeWM applies SIGReg \citep{balestriero2025lejepa} to the encoder embeddings. Let $Z\in\mathbb{R}^{N\times d}$ denote the latent embeddings collected across observations and trajectories. SIGReg samples $M$ unit-norm directions $u^{(m)}\in\mathbb{S}^{d-1}$ and projects the embeddings onto each direction:
\begin{equation}
    h^{(m)} = Zu^{(m)}.
\end{equation}
A univariate Epps-Pulley \citep{epps-pulley} test statistic $\mathcal{T}$ is then applied to each projection,
\begin{equation}
    \mathcal{L}_{\mathrm{SIG}}(Z)
    =
    \frac{1}{M}
    \sum_{m=1}^{M}
    \mathcal{T}\!\left(h^{(m)}\right).
\end{equation}
By encouraging the sampled one-dimensional projections to follow a standard 
Gaussian distribution, SIGReg promotes an approximately isotropic Gaussian 
embedding distribution and prevents the encoder from mapping all inputs to a 
constant.

\begin{wrapfigure}{r}{0.55\linewidth}
\vspace{-1.0\baselineskip}
    \centering
    \includegraphics[width=\linewidth]{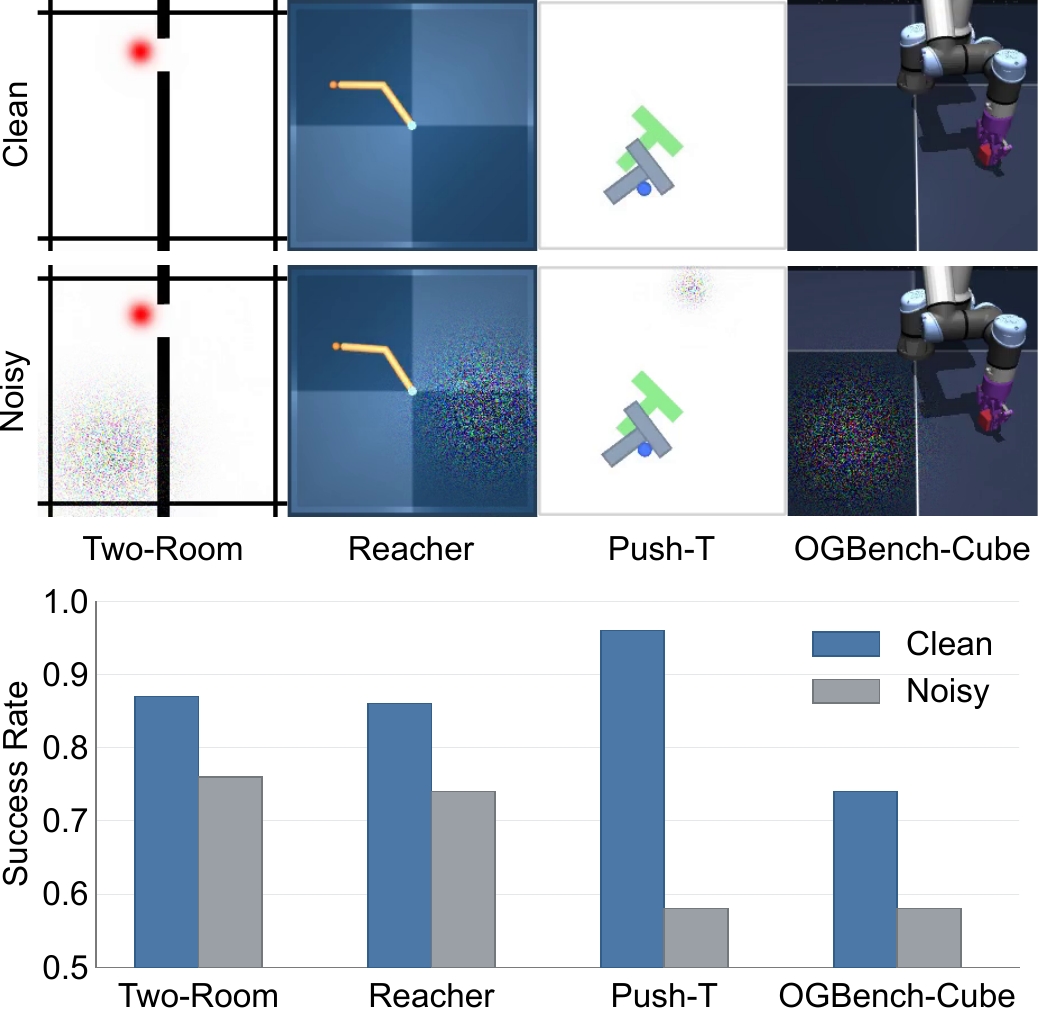}
    \caption{\textbf{Task-irrelevant visual perturbations degrade LeWM's
    latent-space planning.} The $y$-axis starts at 0.5.}
    \label{fig:lewm_noise}
\vspace{-1.0\baselineskip}
\end{wrapfigure}

The representation-level role of SIGReg is important, but distinct 
from constraining the dynamics predictor. In particular, SIGReg regulates the 
geometry of encoder outputs aggregated across samples; it does not directly 
specify how an action-conditioned transition should behave when nuisance 
variation moves an individual observation within that space. If two 
observations correspond to the same physical state but encode to $z_t$ and 
$z_t+\delta$, respectively, non-collapse alone does not require the predictor 
to assign them equivalent future dynamics.


To expose this distinction empirically, we introduce localized Gaussian
noise into task-irrelevant background regions while preserving the
underlying physical state and environment dynamics. We refer to the
resulting observations as \emph{noisy}, in contrast to the original
\emph{clean} observations. The exact perturbation formulation is described in Section~\ref{subsec:exp_setup}.

Figure~\ref{fig:lewm_noise} illustrates this robustness gap: mean planning success with the autoregressive LeWM baseline decreases from $86\%$ to $67\%$ after the background noise is introduced. This motivates examining the
latent-dynamics formulation.

\subsection{Structural Vulnerabilities of Autoregressive Latent Prediction}
Autoregressive predictors construct future trajectories by recursively composing local transitions,
\begin{equation}
    \hat z_{t+k}=g_\phi(\mathcal{H}_{t+k-1},a_{t+k-1}),
    \qquad k=1,\ldots,h,
\end{equation}
where $\mathcal{H}_{t+k-1}$ denotes the latent history within a fixed-size window. This formulation introduces two structural vulnerabilities to perturbations. First, model-generated latent states progressively enter the history used for subsequent transitions, so deviations introduced at early horizons can propagate through the rollout and affect the terminal state used for planning. Second, the one-step objective
\begin{equation}
\mathcal{L}_{\mathrm{pred}}=\|g_\phi(\mathcal{H}_{t},a_t)-z_{t+1}\|_2^2
\end{equation}
directly supervises the predictor at observed latent inputs and clean target endpoints, but does not train it on intermediate perturbed states away from those samples \citep{huang2026vjepa}. If a visual perturbation shifts the encoded observation from $z_t$ to $z_t+\delta$, the transition at the shifted input is not directly specified by the objective. Together, these properties lead to a contrasting formulation that jointly models the future trajectory and trains the dynamics predictor beyond clean latent endpoints.

\section{Method}
\label{sec:method}

\subsection{Training Objective}

We investigate trajectory-level conditional vector-field prediction as an alternative to autoregressive latent dynamics. CFM \citep{lipman2022flow} learns a continuous-time vector field that transports samples from a source distribution $p_0$ toward a target distribution $p_1$. Given a source sample $x_0\sim\mathcal{N}(0,I)$, a target sample $x_1\sim p_{\mathrm{data}}$, and flow time $\tau\sim\mathcal{U}[0,1]$, a commonly used linear path is
\begin{equation}
    x_\tau=(1-\tau)x_0+\tau x_1,
\end{equation}
whose target velocity is constant along the path,
\begin{equation}
    u_\tau(x_\tau\mid x_0,x_1)=x_1-x_0.
\end{equation}
The corresponding CFM objective is
\begin{equation}
    \mathcal{L}_{\mathrm{CFM}}
    =
    \mathbb{E}_{
        \substack{
        x_0 \sim \mathcal{N}(0,I),\\
        x_1 \sim p_{\mathrm{data}},\\
        \tau \sim \mathcal{U}[0,1]
        }}
    \left[
        \left\|
            v_\theta(x_\tau,\tau)
            -
            (x_1-x_0)
        \right\|^2
    \right].
\end{equation}
Training therefore requires only regression of the vector field at randomly sampled intermediate states and does not require numerical ordinary differential equation (ODE) integration.

\begin{figure}[t]
\begin{center}
\includegraphics[width=5.5in]{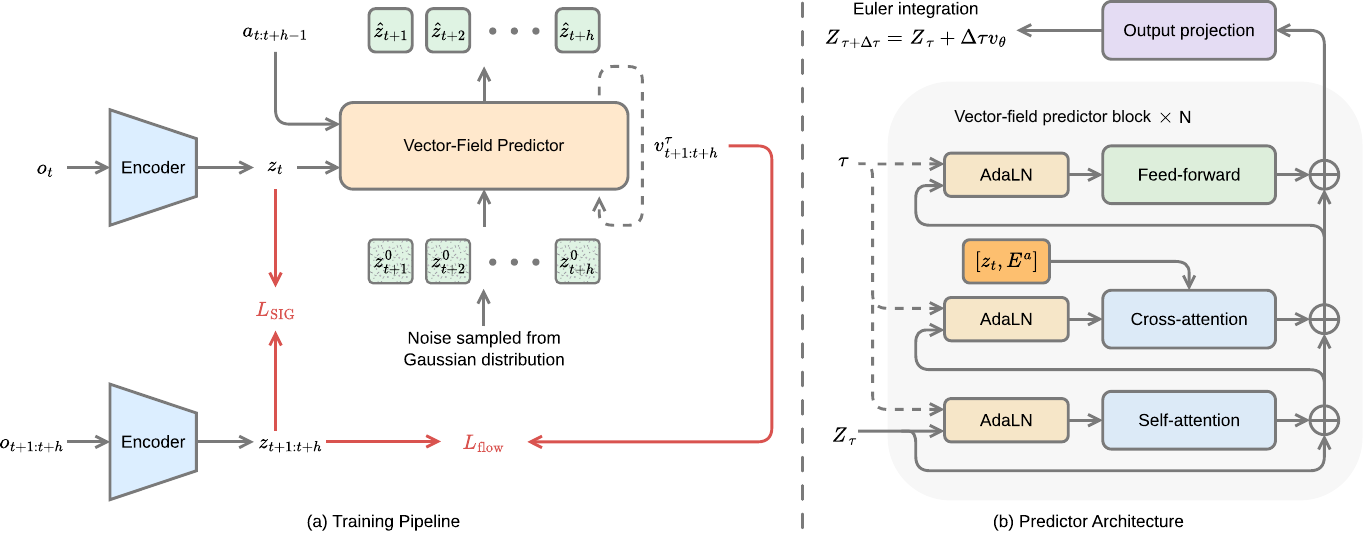}
\end{center}
\caption{\textbf{Overview of F-JEPA.}
\textbf{(a)} training pipeline. The conditional flow matching objective trains the predictor to transport Gaussian source tokens toward future latent representations conditioned on the current latent state and action sequence, while SIGReg regularizes the visual encoder to prevent representation collapse.
\textbf{(b)} predictor architecture. The predictor jointly updates future latent tokens using Transformer blocks with flow-time conditioning and action-dependent context.}
\label{fig:arch}
\end{figure}

We instantiate this formulation for action-conditioned latent world modeling, as illustrated in Figure~\ref{fig:arch}(a). For a prediction horizon of $h$, each training sample contains $h+1$ observations $o_t,o_{t+1},\ldots,o_{t+h}$ and $h$ temporally aligned action blocks $a_t,a_{t+1},\ldots,a_{t+h-1}$. Each action block contains 5 consecutive low-level environment actions, matching the frame skip used in the dataset. A shared visual encoder $f_{\xi}$ maps observations to $d$-dimensional latent states $z_t=f_{\xi}(o_t)\in\mathbb{R}^{d}$,
while an action encoder $h_{\psi}$ maps each action block to the same embedding dimension $e_t^a=h_{\psi}(a_t)\in\mathbb{R}^{d}$.
The future latent trajectory and encoded action sequence are respectively defined as
\begin{equation}
    Z =
    \left[z_{t+1},z_{t+2},\ldots,z_{t+h}\right]
    \in\mathbb{R}^{h\times d},
\end{equation}
\begin{equation}
    E^a =
    \left[e_t^a,e_{t+1}^a,\ldots,e_{t+h-1}^a\right]
    \in\mathbb{R}^{h\times d}.
\end{equation}

For each future position, we independently sample a source latent from a
Gaussian distribution,
\begin{equation}
    z^0_{t+i}
    \sim
    \mathcal{N}(\mu, \sigma^2I),
    \qquad i=1,\ldots,h,
\end{equation}
where $\mu\in\mathbb{R}^d$ and $\sigma>0$ specify the source
distribution. The intermediate trajectory is defined as
\begin{equation}
Z_\tau=(1-\tau)Z_0+\tau Z,
\end{equation}
where $Z_0=[z^0_{t+1},\ldots,z^0_{t+h}]$. Conditioned on the current latent state $z_t$, encoded action sequence $E^a$, and flow time $\tau$, the predictor

\begin{equation}
v_\theta(Z_\tau,\tau,z_t,E^a)
\end{equation}

estimates the velocity field directing the intermediate trajectory toward the clean target trajectory. Following the standard CFM formulation, the training objective is

\begin{equation}
\mathcal{L}_{\mathrm{FM}}
=
\mathbb{E}_{
\substack{
(z_t,E^a,Z)\sim p_{\mathrm{data}},\;
\tau\sim\mathcal{U}[0,1],\\
z^0_{t+i}\overset{\mathrm{i.i.d.}}{\sim}
\mathcal{N}(\mu, \sigma^2I),\; i=1{:}h
}}
\!\left[
\left\|
v_\theta(Z_\tau,\tau,z_t,E^a)-(Z-Z_0)
\right\|^2
\right].
\end{equation}
The two predictor paradigms differ structurally. Autoregressive predictors construct a trajectory by recursively composing one-step transitions, whereas F-JEPA evolves the complete trajectory jointly through a conditional vector field. In addition, autoregressive regression is supervised at observed latent transitions, while flow matching evaluates the vector field along stochastic source--target interpolation paths.

Finally, we adopt SIGReg to prevent representation collapse. The regularizer is applied to the embeddings of both the current and future observations in each sampled sequence. The complete optimization objective is
\begin{equation}
    \mathcal{L}
    =
    \mathcal{L}_{\mathrm{FM}}
    +
    \lambda_{\mathrm{SIG}}
    \mathcal{L}_{\mathrm{SIG}}.
\end{equation}

At inference time, generation starts by drawing a fresh sample from the Gaussian distribution as a source trajectory $Z_0$. Conditioned on $z_t$ and the encoded action sequence $E^a$, the learned vector field defines the ODE
\begin{equation}
\frac{\mathrm{d}Z_\tau}{\mathrm{d}\tau} = v_\theta(Z_\tau,\tau,z_t,E^a).
\end{equation}
We solve this ODE using explicit Euler integration. Given $N_{\text{flow}}$ flow steps, the integration interval is $\Delta\tau=1/N_{\text{flow}}$, and the latent trajectory is updated as
\begin{equation}
    Z_{\tau+\Delta \tau}
    =
    Z_{\tau}
    +
    \Delta\tau\,
    v_\theta(Z_{\tau},\tau,z_t,E^a).
\end{equation}
Starting from $Z_0$, the final integrated sample $Z_{1}$ is taken as the predicted future latent trajectory $\hat Z$.

\subsection{Model Architecture}

Each observation $o_t$ is encoded by a Vision Transformer (ViT) \citep{dosovitskiy2020image}, and its \texttt{[CLS]} representation is projected through a multilayer perceptron (MLP) to obtain the latent state $z_t$. Action blocks are independently mapped to action embeddings by the action encoder. Figure~\ref{fig:arch}(b) illustrates the predictor used by F-JEPA. The dynamics module is a Transformer-based \citep{vaswani2017attention} vector-field predictor. Each future latent token is augmented with a temporal position embedding identifying its prediction horizon. The scalar flow time $\tau$ is first mapped through a sinusoidal embedding and then injected into each Transformer block using adaptive layer normalization (AdaLN) \citep{peebles2023scalable}. The current latent state and encoded action sequence are concatenated as the cross-attention context for future trajectory prediction. Additional implementation details, including the Gaussian source configuration and hyperparameters, are provided in Appendix~\ref{app:implementation}.

\section{Experiments}
\label{sec:exp}

\subsection{Experimental Setup}
\label{subsec:exp_setup}

\paragraph{Planning configurations.}
We evaluate two latent-dynamics formulations on four environments: Two-Room \citep{sobal2025stresstesting}, Reacher \citep{tassa2018deepmind}, Push-T \citep{zhou2024dino}, and OGBench-Cube \citep{park2025ogbench}. See Appendix~\ref{app:b1} for details of the environments and the datasets. We use model-predictive control (MPC) \citep{RICHALET1978413,hansen2022temporal,hansen2024td} with the cross-entropy method (CEM) \citep{cem} for planning. At each planning step, CEM optimizes a sequence of 5 action blocks, each containing 5 low-level actions. Each refinement iteration evaluates 300 randomly sampled candidate sequences and updates the action distribution with the best 30. Only the terminal predicted latent state is used to compute the planning cost. The selected five-block sequence is executed before replanning from the new observation. We report success rates over 50 evaluation rollouts. Planning settings follow LeWM except on Push-T, where F-JEPA uses 10 CEM iterations while the reported LeWM baseline uses its original 30-iteration setting. We examine the effect of this difference explicitly in Section~\ref{sec:ablation}.

\paragraph{Visual perturbations.} To evaluate robustness independently of the underlying task dynamics, we introduce a spatially localized Gaussian perturbation directly in RGB space. Let $p$ denote a pixel location, $k$ an RGB channel, and $c$ the sampled perturbation center. We draw $\epsilon_{p,k}\sim\mathcal{N}(0,s^2)$ and construct
\begin{equation}
    \widetilde{o}_{p,k}
    =
    \operatorname{clip}
    \left(
        o_{p,k}
        +
        \exp\!\left(
            -\frac{\lVert p-c\rVert_2^2}{2r^2}
        \right)
        \epsilon_{p,k},
        0,
        255
    \right),
    \label{eq:visual_noise_patch}
\end{equation}
where $s$ controls the noise magnitude and $r$ controls its spatial decay. The center is sampled from the background, away from the agent and task-relevant objects, and remains fixed within each rollout. In the main comparison, we use $s=100$ and $r=35$ for Two-Room, Reacher, and OGBench-Cube. For Push-T, we use $r=10$, since $r=35$ reduces the LeWM baseline to near-random performance and yields a less informative comparison.

\subsection{Planning Performance under Clean and Perturbed Observations}

\paragraph{Q1: Does trajectory-level vector-field prediction preserve clean-condition performance?}
We first ask whether trajectory-level vector-field prediction compromises planning performance under clean observations relative to autoregressive dynamics. Table~\ref{tab:clean_success_rates} reports numerical results, and Figure~\ref{fig:sr} provides an aggregate comparison with LeWM. F-JEPA matches or exceeds LeWM across all four environments and raises mean success from 86\% to 92\%. The flow-based dynamics formulation therefore preserves strong clean-condition performance.

\begin{table}[h]
\caption{\textbf{Planning success rates (\%) under clean visual conditions.} F-JEPA achieves the highest mean success rate (92\%) and the best or tied-best performance on three of the four environments, showing that the alternative dynamics formulation preserves strong clean-condition planning performance. Results for PLDM \citep{sobal2026learning}, DINO-WM \citep{zhou2024dino}, and LeWM are taken from \citep{maes2026leworldmodel}. The best result in each column is shown in bold.}
\label{tab:clean_success_rates}
\begin{center}
\begin{tabular}{lccccc}
\toprule
\textbf{Method}
& \textbf{Two-Room}
& \textbf{Reacher}
& \textbf{Push-T}
& \textbf{OGBench-Cube}
& \textbf{Mean} \\
\midrule
PLDM     & 97 & 78 & 78 & 65 & 80 \\
DINO-WM  & \textbf{100} & 79 & 74 & \textbf{86} & 85 \\
LeWM     & 87 & 86 & \textbf{96} & 74 & 86 \\
F-JEPA   & \textbf{100} & \textbf{90} & \textbf{96} & 82 & \textbf{92} \\
\bottomrule
\end{tabular}
\end{center}
\end{table}

\paragraph{Q2: How do autoregressive and vector-field dynamics differ under visual perturbations?}
We next evaluate the central question of the paper: whether the alternative latent-dynamics formulation changes sensitivity to task-irrelevant visual variation. Table~\ref{tab:noisy_success_rates} reports performance under localized perturbations that leave the task dynamics unchanged. Because this perturbation evaluation is introduced in our study and corresponding results are unavailable for the other baselines, we focus the robustness comparison on F-JEPA and LeWM, which instantiate vector-field and autoregressive latent dynamics, respectively.

\begin{table}[h]
\caption{\textbf{Planning success rates (\%) under noisy visual conditions.}
F-JEPA improves performance across all four environments and increases the mean success rate from 67\% to 86\%. The final row reports the absolute percentage-point difference from LeWM.}
\label{tab:noisy_success_rates}
\begin{center}
\begin{tabular}{lccccc}
\toprule
\textbf{Method}
& \textbf{Two-Room}
& \textbf{Reacher}
& \textbf{Push-T}
& \textbf{OGBench-Cube}
& \textbf{Mean} \\
\midrule
LeWM     & 76 & 74 & 58 & 58 & 67 \\
F-JEPA   & \textbf{96} & \textbf{90} & \textbf{90} & \textbf{66} & \textbf{86} \\
$\Delta$ & +20 & +16 & +32 & +8 & +19 \\
\bottomrule
\end{tabular}
\end{center}
\end{table}

\begin{figure}[ht]
\begin{center}
\includegraphics[width=5.5in]{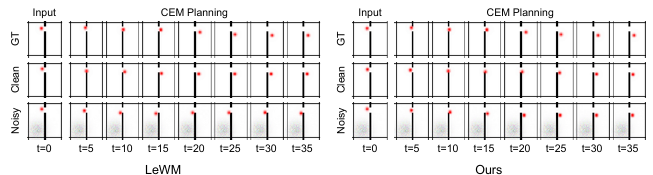}
\end{center}
\caption{\textbf{Qualitative rollouts in Two-Room.}
Both LeWM (left) and F-JEPA (right) reach the goal under clean observations,
whereas LeWM fails after task-irrelevant background noise is introduced
and F-JEPA remains successful, illustrating improved robustness to
visual perturbations.}
\label{fig:tworoom_rollouts}
\end{figure}

\begin{figure}[ht]
\begin{center}
\includegraphics[width=5.5in]{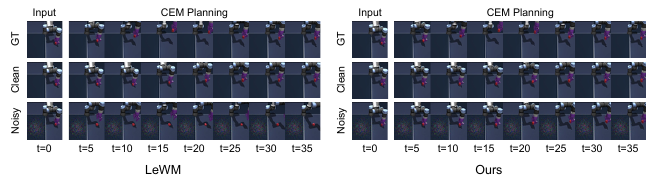}
\end{center}
\caption{\textbf{Qualitative rollouts in OGBench-Cube.}
Both LeWM (left) and F-JEPA (right) complete the task under clean observations,
whereas LeWM fails after task-irrelevant background noise is introduced
and F-JEPA remains successful, illustrating improved robustness to
visual perturbations.}
\label{fig:cube_rollouts}
\end{figure}

F-JEPA improves success in every environment under visual perturbations, raising the mean from 67\% to 86\%, a difference of 19 percentage points. The mean degradation from clean to noisy observations is 19 points for LeWM and 6 points for F-JEPA. This substantially smaller degradation suggests that the choice between recursive autoregressive prediction and joint vector-field evolution can substantially affect planning robustness. These results are consistent with the hypothesis that stochastic trajectory-level training helps the predictor tolerate latent deviations.

The qualitative rollouts in Figures~\ref{fig:tworoom_rollouts} and~\ref{fig:cube_rollouts} illustrate the same trend. Both methods succeed under clean observations, while after task-irrelevant visual perturbations are introduced, LeWM fails in both examples whereas F-JEPA remains successful.

\subsection{Perturbation and Inference Stress Tests}
\label{sec:ablation}

\begin{wraptable}{r}{0.55\textwidth}
\vspace{-2.0\baselineskip}
\centering
\caption{\textbf{Reacher success rates (\%) across Gaussian-patch radii.}
F-JEPA consistently outperforms LeWM across all tested perturbation radii, indicating that its robustness gain is not specific to a particular spatial noise scale. The noise standard deviation is fixed at $s=100$, and $\Delta$ denotes the absolute percentage-point improvement over LeWM.}
\label{tab:radius_ablation}
\vspace{4pt}
\resizebox{0.98\linewidth}{!}{%
\begin{tabular}{lccccccc}
\toprule
& \multicolumn{7}{c}{\textbf{Gaussian-patch radius $r$ (pixels)}} \\
\cmidrule(lr){2-8}
\textbf{Method} & 10 & 15 & 20 & 25 & 30 & 35 & 40 \\
\midrule
LeWM   & 80 & 84 & 72 & 74 & 68 & 74 & 72 \\
F-JEPA & \textbf{88} & \textbf{86} & \textbf{82} & \textbf{88}
       & \textbf{82} & \textbf{90} & \textbf{84} \\
$\Delta$ & +8 & +2 & +10 & +14 & +14 & +16 & +12 \\
\bottomrule
\end{tabular}
}
\end{wraptable}

\paragraph{Q3: Does the robustness advantage persist across perturbation and inference settings?}
We examine whether the observed robustness difference is confined to a particular perturbation configuration or inference budget. We vary perturbation radius and magnitude, flow integration steps, and CEM refinement iterations; additional results across environments are provided in Appendix~\ref{app:b2}.

\noindent\textbf{Robustness across perturbation radius.}
We first vary the spatial extent of the perturbation on Reacher while holding all other settings fixed. Table~\ref{tab:radius_ablation} shows that F-JEPA consistently outperforms LeWM across tested radii, so the robustness gain is not specific to a single spatial perturbation scale.

\begin{table}[t]
\begin{minipage}[t]{0.49\textwidth}
\centering
\caption{\textbf{Reacher success rates (\%) across Gaussian-noise standard
deviations.} F-JEPA consistently outperforms LeWM across all tested noise
levels and maintains an advantage even at the largest perturbation magnitude.
The patch radius is fixed at $r=35$, and $\Delta$ denotes the absolute
percentage-point improvement over LeWM.}
\label{tab:std_ablation}
\vspace{4pt}

\footnotesize
\setlength{\tabcolsep}{3pt}

\begin{tabularx}{\linewidth}{
@{\hspace{3pt}}
>{\raggedright\arraybackslash}p{0.22\linewidth}
*{4}{>{\centering\arraybackslash}X}
@{\hspace{3pt}}
}
\toprule
& \multicolumn{4}{c}{\textbf{Noise standard deviation $s$}} \\
\cmidrule(lr){2-5}
\textbf{Method} & 50 & 100 & 150 & 200 \\
\midrule
LeWM    & 82 & 74 & 66 & 56 \\
F-JEPA  & \textbf{88} & \textbf{90} & \textbf{78} & \textbf{76} \\
$\Delta$ & +6 & +16 & +12 & +20 \\
\bottomrule
\end{tabularx}
\end{minipage}
\hfill
\begin{minipage}[t]{0.49\textwidth}
\centering
\caption{\textbf{Push-T success rates (\%) across flow integration steps and
corresponding inference cost (s).} 8 Euler steps achieve the highest success
rates. Times report wall-clock seconds over 50 rollouts; bold denotes the
highest success rate within each condition.}
\label{tab:flow_steps_ablation}
\vspace{7pt}

\footnotesize
\setlength{\tabcolsep}{3pt}

\begin{tabularx}{\linewidth}{
@{\hspace{3pt}}
>{\centering\arraybackslash}p{0.22\linewidth}
*{4}{>{\centering\arraybackslash}X}
@{\hspace{3pt}}
}
\toprule
&
\multicolumn{2}{c}{\textbf{Clean}} &
\multicolumn{2}{c}{\textbf{Noisy}} \\
\cmidrule(lr){2-3}\cmidrule(lr){4-5}

\textbf{Flow steps} & \textbf{Success} & \textbf{Time}
& \textbf{Success} & \textbf{Time} \\
\midrule

4  & \textbf{96} & 31.7  & 82          & 35.0  \\
8  & \textbf{96} & 42.7  & \textbf{90} & 47.7  \\
16 & \textbf{96} & 66.9  & 82          & 78.0  \\
32 & 92          & 111.8 & 84          & 130.2 \\

\bottomrule
\end{tabularx}
\end{minipage}
\end{table}

\paragraph{Robustness across perturbation magnitude.}
We next hold the patch radius fixed and vary the noise standard deviation. Table~\ref{tab:std_ablation} shows that F-JEPA remains stronger across every tested magnitude, including the largest perturbation. Together, the radius and magnitude sweeps show that the observed robustness difference is not confined to a single perturbation configuration.

\begin{wraptable}{r}{0.45\textwidth}
\vspace{-2.0\baselineskip}
\centering
\caption{\textbf{Push-T success rates (\%) across CEM iterations.}
F-JEPA maintains strong clean performance across all tested planning budgets and achieves its best noisy success rate with only 10 CEM iterations. Under the same 10-iteration budget, F-JEPA substantially outperforms LeWM under noise. Bold denotes the highest success rate for each method and visual condition.}
\label{tab:cem_steps_ablation}
\vspace{4pt}
\resizebox{0.98\linewidth}{!}{%
\begin{tabular}{lcccccc}
\toprule
&
\multicolumn{3}{c}{\textbf{Clean}} &
\multicolumn{3}{c}{\textbf{Noisy}} \\
\cmidrule(lr){2-4}\cmidrule(lr){5-7}
\textbf{Method} & 10 & 20 & 30 & 10 & 20 & 30 \\
\midrule
LeWM   & 94 & 92 & \textbf{96}
       & \textbf{60} & 56 & 58 \\
F-JEPA & \textbf{96} & \textbf{96} & \textbf{96}
       & \textbf{90} & 88 & 84 \\
\bottomrule
\end{tabular}
}
\end{wraptable}

\noindent\textbf{Sensitivity to flow integration.}
Table~\ref{tab:flow_steps_ablation} studies the accuracy--computation
trade-off of ODE integration on Push-T. 8 Euler steps preserve the
best clean performance while achieving the highest noisy success rate.
Increasing the number of integration steps beyond 8 substantially
increases runtime without improving control performance, indicating
that finer numerical integration is not the limiting factor in this
setting. We therefore use 8 flow steps in the main experiments.

\noindent\textbf{Sensitivity to planning budget.}
Finally, we examine whether the Push-T improvement can be attributed to differences in CEM planning compute. The main F-JEPA result uses 10 CEM iterations, whereas the reported LeWM baseline retains its original 30-iteration setting. Table~\ref{tab:cem_steps_ablation} evaluates both methods under matched CEM budgets. F-JEPA maintains 96\% clean success across the full sweep and achieves its highest noisy success with only 10 iterations. Additional robustness and inference analyses are provided in Appendix~\ref{app:b2}.

\section{Conclusion}
\label{sec:conclusion}

We studied how latent-dynamics formulation affects planning robustness in
reconstruction-free JEPA world models, focusing on the contrast
between autoregressive state-transition prediction and
trajectory-level conditional vector-field prediction. Across four
control environments, the vector-field formulation substantially
reduces degradation under task-irrelevant visual perturbations compared with 
autoregressive prediction while maintaining strong
performance under clean observations. This advantage persists across
perturbation strengths and inference settings,
highlighting flow-based latent dynamics as a promising alternative to deterministic autoregressive prediction for robust JEPA world modeling.

\textbf{Limitations and future work.}
Our comparison does not isolate the individual contributions of joint
trajectory prediction and stochastic flow training, or their interaction. 
Controlled intermediate formulations remain important
future work. Our evaluation is also limited to relatively short-horizon
tasks and primarily localized Gaussian perturbations, motivating studies
of longer horizons and broader distribution shifts. Finally, flow
inference requires multiple vector-field evaluations, making more
efficient flow-based latent prediction another important direction.

\subsection*{AI use statement}

In this work, we used generative AI tools to assist with formulating mathematical claims, providing feedback on experiments, implementing methods, and interpreting results. We did not use generative AI tools to develop theoretical models or conceptual frameworks or to propose or refine hypotheses. The following uses were not applicable to this work: generating synthetic datasets, providing critical ingredients for proving mathematical claims, assisting in the writing of proofs, assisting with translation, cleaning or reformatting datasets, and supporting qualitative or thematic data analysis.

Additionally, we used generative AI tools to edit the research paper for improved readability, identify relevant literature, and propose a title for the paper. We reviewed all AI-assisted work. LLM-generated code was verified and tested for correctness by the authors, and LLM-generated mathematical claims were examined and refined to ensure that they were correct and aligned with the actual method. We take responsibility for the final content of this work, including all text, claims, and artifacts produced with the aid of generative AI.

\subsection*{Reproducibility statement}

For reproducibility, Section~\ref{sec:method} provides the complete training objective and model architecture. Appendix~\ref{app:implementation} documents implementation details, training hyperparameters, Gaussian source configurations, and hardware used for all experiments. The planning and evaluation protocols are specified in Section~\ref{subsec:exp_setup}, and test environments and datasets are detailed in Appendix~\ref{app:b1}. We also analyze key inference-time parameters, including flow integration steps and CEM refinement iterations, to make the reported performance and computational settings directly reproducible.

\bibliography{iclr2027_conference}
\bibliographystyle{iclr2027_conference}

\clearpage
\appendix
\section{Implementation Details}
\label{app:implementation}

\paragraph{Flow source and attention configuration.} We consider two Gaussian source parameterizations. The \emph{standard-noise} source independently samples each future token as $z^0_{t+i}\sim\mathcal{N}(0,I)$, whereas the \emph{state-centered} source samples $z^0_{t+i}\sim\mathcal{N}(z_t,0.5^2 I)$. We further consider either bidirectional or causal self-attention among future trajectory tokens. Under causal self-attention, the token at future position $i$ attends only to itself and preceding future positions, whereas bidirectional self-attention allows interactions across the complete predicted trajectory. In both variants, action conditioning remains temporally causal: the token at position $i$ receives only the action prefix $a_{t:t+i-1}$. Two-Room and OGBench-Cube use the standard-noise source with bidirectional future-token self-attention, while Reacher and Push-T use the state-centered source with causal future-token self-attention.

\paragraph{Architecture.} 
All observations are resized to $224\times224$ and normalized before being passed to the visual encoder. We use a ViT-Tiny with $14\times14$ patches as the visual encoder and an MLP as the action encoder. Both projected visual latents and action embeddings have dimension 192. The visual encoder and latent dimensionality are matched between the autoregressive and vector-field models to enable a controlled comparison of the dynamics formulations. For the vector-field predictor, we use 6 Transformer blocks with 16 attention heads, head dimension 64, and an MLP hidden dimension of 2048.

\paragraph{Training details.} All F-JEPA models are trained for 20 epochs with a batch size of 128, a prediction horizon of 5, and a history size of 1. For Push-T, the selected model initializes the visual encoder and projector from a trained checkpoint with autoregressive formulation and keeps both components frozen during training; predictor dropout is set to 0. For Reacher, the visual encoder and projector are also initialized from a checkpoint with autoregressive formulation but remain trainable as a warm start. The remaining selected models learn the visual representation jointly from training and use a predictor dropout of 0.1. The SIGReg weight is set to 0.09 for Push-T and 0.1 for the remaining environments. All other preprocessing, optimization, and training settings follow LeWM.

\paragraph{Hardware.} All training and evaluation experiments are conducted on a single NVIDIA RTX 6000 Ada GPU.

\section{Datasets and Additional Results}
\label{app:b}

\subsection{Test Environments and Datasets}
\label{app:b1}
\begin{figure}[ht]
\begin{center}
\includegraphics[width=4.0in]{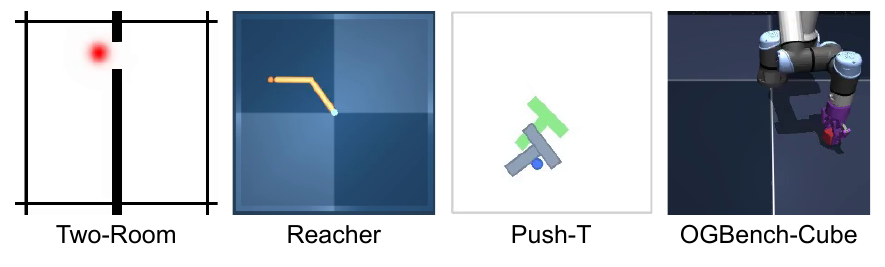}
\end{center}
\caption{\textbf{Test environments.} Representative observations from the four evaluation environments: Two-Room, Reacher, Push-T, and OGBench-Cube, covering navigation, continuous control, and object manipulation tasks.}
\label{fig:envs}
\end{figure}

We evaluate on four visually distinct control environments covering navigation and manipulation, illustrated in Figure~\ref{fig:envs}: Two-Room, a goal-directed navigation task in a two-room layout; Reacher, a continuous-control task requiring a planar arm to reach a target; Push-T, a planar manipulation task in which an agent pushes a T-shaped object toward a goal configuration; and OGBench-Cube, a goal-conditioned cube manipulation task.

We use the same dataset construction as the autoregressive baseline. Observations are recorded with a frame skip of 5, such that two consecutive observation frames are separated by 5 low-level environment steps. The corresponding low-level actions are concatenated into an action block, containing the 5 actions executed between adjacent observations. Consequently, each latent transition in the world model is conditioned on one action block rather than a single low-level action.

\subsection{Additional Robustness and Inference Analysis}
\label{app:b2}

\begin{figure}[ht]
\begin{center}
\includegraphics[width=5.5in]{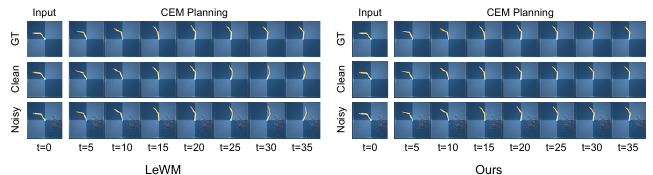}
\end{center}
\caption{\textbf{Qualitative rollouts in Reacher.}
LeWM (left) and F-JEPA (right) are shown under clean and noisy
observations. Under clean observations, both methods complete the task,
while F-JEPA reaches the target with higher precision. After visual
perturbations are introduced, LeWM fails to reach the target, whereas
F-JEPA remains successful.}
\label{fig:reacher_rollouts}
\end{figure}

\begin{figure}[ht]
\begin{center}
\includegraphics[width=5.5in]{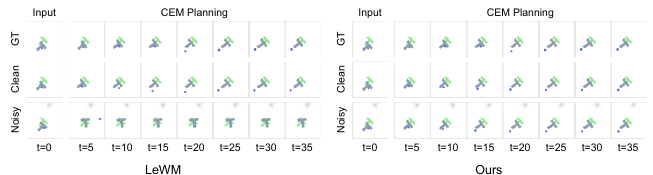}
\end{center}
\caption{\textbf{Qualitative rollouts in Push-T.}
LeWM (left) and F-JEPA (right) are shown under clean and noisy
observations. Both methods complete the task under clean observations,
whereas F-JEPA remains successful under visual perturbations while LeWM
fails.}
\label{fig:pusht_rollouts}
\end{figure}

\paragraph{Qualitative rollouts.} Figures~\ref{fig:reacher_rollouts} and~\ref{fig:pusht_rollouts} provide additional comparisons between LeWM and F-JEPA. Consistent with the main-text examples, F-JEPA remains more reliable after visual perturbations are introduced. In Reacher, both methods complete the task under the clean condition, but F-JEPA reaches the target with higher precision. Together, these rollouts support the improved planning performance and robustness observed across the four environments.

\begin{table}[ht]
\caption{\textbf{Additional analysis across Gaussian-patch radii.} Complementing the Reacher results in the main text, F-JEPA matches or outperforms LeWM across all tested radii in Two-Room, Push-T, and OGBench-Cube, indicating that its robustness advantage persists across different spatial perturbation scales. The noise standard deviation is fixed at $s=100$, and $\Delta$ denotes the absolute percentage-point difference between F-JEPA and LeWM.}
\label{tab:appendix_radius_all}
\begin{center}
\small
\setlength{\tabcolsep}{4pt}
\begin{tabular}{llccccccc}
\toprule
&& \multicolumn{7}{c}{\textbf{Gaussian-patch radius $r$ (pixels)}} \\
\cmidrule(lr){3-9}
\textbf{Environment} & \textbf{Method} & 10 & 15 & 20 & 25 & 30 & 35 & 40 \\
\midrule
Two-Room
& LeWM   & 78 & 76 & 80 & 82 & 78 & 76 & 72 \\
& F-JEPA & \textbf{100} & \textbf{100} & \textbf{98} & \textbf{98}
         & \textbf{92} & \textbf{96} & \textbf{80} \\
& $\Delta$ & +22 & +24 & +18 & +16 & +14 & +20 & +8 \\
\addlinespace
Push-T
& LeWM   & 58 & 42 & 32 & 22 & 6 & \textbf{8} & 4 \\
& F-JEPA & \textbf{90} & \textbf{80} & \textbf{40} & \textbf{26}
         & \textbf{22} & \textbf{8} & \textbf{14} \\
& $\Delta$ & +32 & +38 & +8 & +4 & +16 & 0 & +10 \\
\addlinespace
OGBench-Cube
& LeWM   & 58 & 60 & 60 & 56 & 54 & 58 & 44 \\
& F-JEPA & \textbf{68} & \textbf{70} & \textbf{64} & \textbf{58}
         & \textbf{60} & \textbf{66} & \textbf{54} \\
& $\Delta$ & +10 & +10 & +4 & +2 & +6 & +8 & +10 \\
\bottomrule
\end{tabular}
\end{center}
\end{table}

\begin{figure}[ht]
\begin{center}
\includegraphics[width=5.5in]{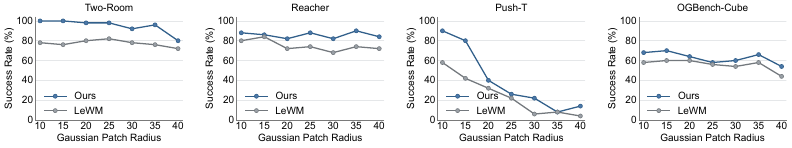}
\end{center}
\caption{\textbf{Success-rate trends across Gaussian-patch radii.}
F-JEPA generally maintains higher success rates than LeWM across
different spatial perturbation scales in all four environments,
complementing the quantitative results in
Tables~\ref{tab:radius_ablation} and~\ref{tab:appendix_radius_all}.}
\label{fig:radius_ablation_all}
\end{figure}

\paragraph{Robustness across perturbation severity.} To complement the Reacher perturbation analysis reported in the main text, we evaluate the remaining environments across Gaussian-patch radii and noise magnitudes. These results test whether the robustness trends observed in the main experiments persist across different perturbation scales and environments.

\begin{table}[ht]
\caption{\textbf{Additional analysis across Gaussian-noise magnitudes.} Across Two-Room, Push-T, and OGBench-Cube, F-JEPA generally maintains higher success than LeWM as the perturbation magnitude increases, showing that the observed robustness difference is not confined to the noise level used in the main evaluation. The patch radius is fixed at $r=35$ for Two-Room and OGBench-Cube and at $r=10$ for Push-T, and $\Delta$ denotes the absolute percentage-point difference between F-JEPA and LeWM.}
\label{tab:appendix_std_all}
\begin{center}
\begin{tabular}{llcccc}
\toprule
&& \multicolumn{4}{c}{\textbf{Noise standard deviation $s$}} \\
\cmidrule(lr){3-6}
\textbf{Environment} & \textbf{Method} & 50 & 100 & 150 & 200 \\
\midrule
Two-Room
& LeWM   & 76 & 76 & 70 & 68 \\
& F-JEPA & \textbf{100} & \textbf{96} & \textbf{84} & \textbf{80} \\
& $\Delta$ & +24 & +20 & +14 & +12 \\
\addlinespace
Push-T
& LeWM   & 86 & 58 & 56 & 54 \\
& F-JEPA & \textbf{94} & \textbf{90} & \textbf{72} & \textbf{72} \\
& $\Delta$ & +8 & +32 & +16 & +18 \\
\addlinespace
OGBench-Cube
& LeWM   & 68 & 58 & \textbf{62} & 60 \\
& F-JEPA & \textbf{72} & \textbf{66} & 60 & \textbf{62} \\
& $\Delta$ & +4 & +8 & -2 & +2 \\
\bottomrule
\end{tabular}
\end{center}
\end{table}

\begin{figure}[ht]
\begin{center}
\includegraphics[width=5.5in]{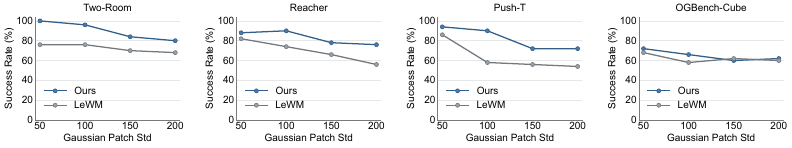}
\end{center}
\caption{\textbf{Success-rate trends across Gaussian-noise magnitudes.}
F-JEPA generally retains a performance advantage over LeWM as the noise
standard deviation increases, indicating improved robustness across a
broad range of perturbation strengths. Corresponding numerical results
are reported in Tables~\ref{tab:std_ablation} and~\ref{tab:appendix_std_all}.}
\label{fig:std_ablation_all}
\end{figure}

Tables~\ref{tab:appendix_radius_all} and~\ref{tab:appendix_std_all} show that the robustness difference is not tied to a single perturbation configuration. Across the radius sweep, F-JEPA matches or exceeds LeWM at every tested setting in the three environments. The noise-magnitude sweep exhibits a similar overall trend: F-JEPA remains stronger across nearly all settings, including the largest tested perturbations, with only a small reversal on OGBench-Cube at $s=150$. Push-T becomes particularly challenging as the perturbation grows spatially, yet F-JEPA retains an advantage or matches the baseline throughout the sweep. Together with the Reacher results in the main text, these experiments show that the observed difference persists across a broad range of visual perturbation strengths and spatial scales. Figures~\ref{fig:radius_ablation_all} and~\ref{fig:std_ablation_all} summarize the corresponding success-rate trends across all four environments.

\paragraph{Sensitivity to flow integration.} We additionally evaluate the effect of ODE integration steps on Two-Room, Reacher, and OGBench-Cube, complementing the Push-T analysis reported in the main text.

\begin{figure}[ht]
\begin{center}
\includegraphics[width=5.5in]{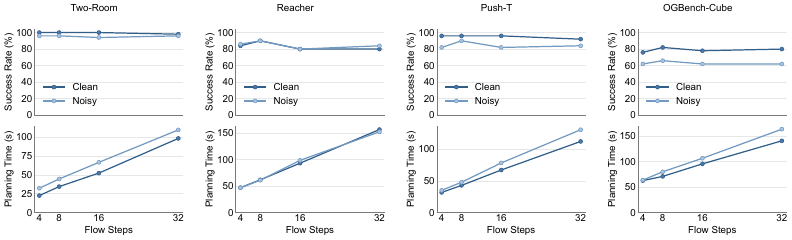}
\end{center}
\caption{\textbf{Planning performance and inference time across flow
integration steps.}
Success rates under clean and noisy observations and corresponding
planning times are shown across different Euler integration budgets for
all four environments. Increasing the number of flow steps does not
consistently improve planning performance, while inference time grows
substantially with additional integration steps. Overall, 8 steps provide
a favorable performance-efficiency trade-off across the evaluated tasks. Corresponding numerical results are reported in Tables~\ref{tab:flow_steps_ablation} and~\ref{tab:appendix_flow_steps_all}.}
\label{fig:flow_steps_ablation_all}
\end{figure}

\begin{table}[ht]
\caption{\textbf{Additional analysis across flow integration steps.} Complementing the Push-T results in the main text, 8 Euler steps achieve the best or tied-best planning performance across the remaining environments, while additional integration steps substantially increase runtime without yielding consistent performance gains. Times report wall-clock seconds over 50 rollouts.}
\label{tab:appendix_flow_steps_all}
\begin{center}
\small
\begin{tabular}{lccccc}
\toprule
&& \multicolumn{2}{c}{\textbf{Clean}} &
\multicolumn{2}{c}{\textbf{Noisy}} \\
\cmidrule(lr){3-4}\cmidrule(lr){5-6}
\textbf{Environment} & \textbf{Flow steps}
& \textbf{Success (\%)} & \textbf{Time (s)}
& \textbf{Success (\%)} & \textbf{Time (s)} \\
\midrule
Two-Room
& 4  & \textbf{100} & 22.5 & \textbf{96} & 32.2 \\
& 8  & \textbf{100} & 34.4 & \textbf{96} & 44.5 \\
& 16 & \textbf{100} & 52.3 & 94          & 66.5 \\
& 32 & 98            & 98.2 & \textbf{96} & 109.5 \\
\addlinespace
Reacher
& 4  & 84          & 47.1  & 86          & 47.0 \\
& 8  & \textbf{90} & 62.0  & \textbf{90} & 61.2 \\
& 16 & 80          & 93.4  & 80          & 98.8 \\
& 32 & 80          & 157.1 & 84          & 152.8 \\
\addlinespace
OGBench-Cube
& 4  & 76          & 62.8  & 62          & 63.8 \\
& 8  & \textbf{82} & 71.1  & \textbf{66} & 80.2 \\
& 16 & 78          & 95.7  & 62          & 106.7 \\
& 32 & 80          & 141.0 & 62          & 163.9 \\
\bottomrule
\end{tabular}
\end{center}
\end{table}

Table~\ref{tab:appendix_flow_steps_all} confirms the trend observed on Push-T. Increasing the number of Euler steps does not monotonically improve planning performance: Two-Room is already near saturation with 4--8 steps, while both Reacher and OGBench-Cube achieve their strongest clean and noisy results with 8 steps. In contrast, inference time grows substantially as the number of integration steps increases. These results support 8 Euler steps as a favorable accuracy-efficiency trade-off across the evaluated environments and indicate that finer numerical integration is not the primary bottleneck for planning performance. Figure~\ref{fig:flow_steps_ablation_all} summarizes the effect of the
number of Euler integration steps across all four environments.

\paragraph{Inference-time analysis.} We compare the inference cost of F-JEPA and LeWM on Two-Room using the same 10-iteration CEM planning budget. F-JEPA requires multiple vector-field evaluations to integrate the flow trajectory, whereas LeWM autoregressively applies its one-step predictor across the 5-step prediction horizon. Table~\ref{tab:inference_time} reports both the latency of the dynamics model and the resulting end-to-end evaluation time. 

\begin{table}[t] 
\caption{\textbf{Inference-time comparison between LeWM and F-JEPA.} F-JEPA incurs higher dynamics and per-planning latency due to iterative flow integration, while the end-to-end runtime gap over 50 rollouts is smaller; over these rollouts, LeWM requires substantially more replanning during task execution. Both methods use 10 CEM refinement iterations.} 
\label{tab:inference_time} 
\begin{center} \small 
\begin{tabular}{lccc} 
\toprule 
\textbf{Method} & \textbf{Inference latency (ms)} & \textbf{Planning time (ms)} & \textbf{50 rollouts (s)} \\ 
\midrule 
LeWM & 2.1 / prediction & 105.0 & 21.7 \\ F-JEPA & 5.0 / flow step & 400.0 & 34.4 \\ 
\bottomrule 
\end{tabular} 
\end{center} 
\end{table} 

A single flow step is more expensive than a one-step LeWM prediction (5.0 ms versus 2.1 ms). Moreover, one F-JEPA planning cycle performs 8 flow integration steps over 10 CEM iterations, resulting in 400.0 ms per planning step, compared with 105.0 ms for the 5 autoregressive predictions used by LeWM. Thus, F-JEPA is approximately $3.8\times$ slower per planning cycle. However, this difference does not translate directly to end-to-end evaluation time: over 50 rollouts, F-JEPA takes 34.4 s compared with 21.7 s for LeWM, corresponding to a $1.6\times$ increase. This smaller practical gap arises because LeWM requires substantially more replanning steps over complete rollouts, whereas F-JEPA more often completes the task with fewer replanning cycles. These results highlight that the additional cost of iterative flow integration is partially offset at the task level by more effective planning. Moreover, because F-JEPA updates the complete future trajectory jointly at each flow step rather than recursively applying a one-step predictor, its relative computational trade-off may become more favorable for longer prediction horizons, where autoregressive rollout requires an increasing number of sequential model evaluations.

\end{document}